\documentclass{article}
\usepackage{spconf,amsmath,graphicx,amssymb,bm}

\title{Attention-Based End-to-End Speech Recognition on voice search}
\name{\em Changhao Shan$^{1, 2}$, Junbo Zhang$^2$, Yujun Wang$^2$, Lei Xie$^1$}
\address{$^1$Shaanxi Provincial Key Laboratory of Speech and Image Information Processing, \\
	School of Computer Science, Northwestern Polytechnical University, Xi'an, China \\
	$^2$Xiaomi Inc., Beijing, China \\
	{\small \tt {\{shanchanghao, zhangjunbo, wangyujun\}}@xiaomi.com, {lxie}@nwpu.edu.cn}
}
\begin{document}
%
  \maketitle
  \begin{abstract}
    Recently, there has been a growing interest in end-to-end speech recognition that directly transcribes speech to text without any predefined alignments. In this paper, we explore the use of attention-based encoder-decoder model for Mandarin speech recognition on a voice search task. Previous attempts have shown that applying attention-based encoder-decoder to Mandarin speech recognition was quite difficult due to the logographic orthography of Mandarin, the large vocabulary and the conditional dependency of the attention model. In this paper, we use character embedding to deal with the large vocabulary. Several tricks are used for effective model training, including L2 regularization, Gaussian weight noise and frame skipping. We compare two attention mechanisms and use attention smoothing to cover long context in the attention model. Taken together, these tricks allow us to finally achieve a character error rate (CER) of 3.58\% and a sentence error rate (SER) of 7.43\% on the MiTV voice search dataset. While together with a trigram language model, CER and SER reach 2.81\% and 5.77\%, respectively.
  \end{abstract}

\vspace{-5pt}
  \section{Introduction}\vspace{-5pt}
    Voice search (VS) allows users to acquire information by a simple voice command. It has become a dominating function on various devices such as smart phones, speakers and TVs, etc. Automatic speech recognition (ASR) is the first step for a voice search task and thus its performance highly affects the user experience.
    
    Deep Neural Networks (DNNs) have shown tremendous success and are widely used in ASR, usually in combination with Hidden Markov Models (HMMs)~\cite{hinton2012deep, deng2013recent, schmidhuber2015deep}. These systems are based on a complicated architecture with several separate components, including acoustic, phonetic and language models, which are usually trained separately, each with a different objective. Recently, some \textit{end-to-end} neural network ASR approaches, such as connectionist temporal classification (CTC)~\cite{graves2006connectionist, amodei2016deep, miao2015eesen} and attention-based encoder-decoder~\cite{chorowski2014end, chorowski2015attention, bahdanau2016end, chan2016listen, chan2016online, Hori2017Advances, zhang2017very, chorowski2016towards}, have emerged. These end-to-end trained systems directly map the input acoustic speech to grapheme (or word) sequences and the acoustic, pronunciation, and language modeling components are trained jointly in a single system.
    

Attention-based models have become increasingly popular and with delightful performances on various sequence-to-sequence tasks, such as machine translation~\cite{bahdanau2014neural}, text summarization~\cite{rush2015neural}, image captioning~\cite{xu2015show} and speech recognition. 
    In speech recognition, the attention-based approaches usually consist of an encoder network, which maps the input acoustic speech into a higher-level representation, and an attention-based decoder that predicts the next output symbol conditioned on the sequence of previous predictions.  A recent comparison of sequence-to-sequence models for speech recognition~\cite{Prabhavalkar2017} has shown that Listen, Attend and Spell (LAS)~\cite{chan2016listen}, a typical attention-based approach, offered improvements over other sequence-to-sequence models.

    Attention-based encoder-decoder performs considerably well in English speech recognition~\cite{chorowski2016towards} and many attempts have been proposed to further optimize the model~\cite{chorowski2015attention,zhang2017very}.
    However, applying attention-based encoder-decoder to Mandarin was found quite problematic. In~\cite{chan2016online}, Chan \textit{et. al.} have pointed out that the attention model is difficult to converge with Mandarin data due to the logographic orthography of Mandarin, the large vocabulary and the conditional dependency of the attention model. They have proposed a joint Mandarin Character-Pinyin model but with limited success: the character error rate (CER) is as high as 59.3\%  on GALE  broadcast news corpus. In this paper, we aim to improve the LAS approach for Mandarin speech recognition on a voice search task.  Instead of using joint Character-Pinyin model, we directly use Chinese characters as network output. Specifically, we map the one-hot character representation to an embedding vector via a neural network layer. We also use several tricks for effective model training, including L2 regularization~\cite{hinton1993keeping}, Gaussian weight noise~\cite{jim1996analysis} and frame skipping~\cite{miao2016simplifying}. We compare two attention mechanisms and use attention smoothing to cover long context in the attention model. Taken together, these tricks allow us to finally achieve a promising result on a Mandarin voice search task.
	


  \section{Listen, Attend and Spell}
    Listen, Attend and Spell (LAS)~\cite{chan2016listen} is an attention-based encoder-decoder network which is often used to deal with variable-length input to variable-length output mapping problems. The encoder (the Listen module) extracts a higher-level feature representation (i.e., an embedding) from the input features. Then the attention mechanism (the Attend module) determines which encoder features should be attended in order to predict the next output symbol, resulting in a context vector. Finally, the decoder (the Spell module) takes the attention context vector and an embedding of the previous prediction to generate a prediction of the next output.
    
\begin{figure}[t]
    	\centering
    	\includegraphics[width=\columnwidth]{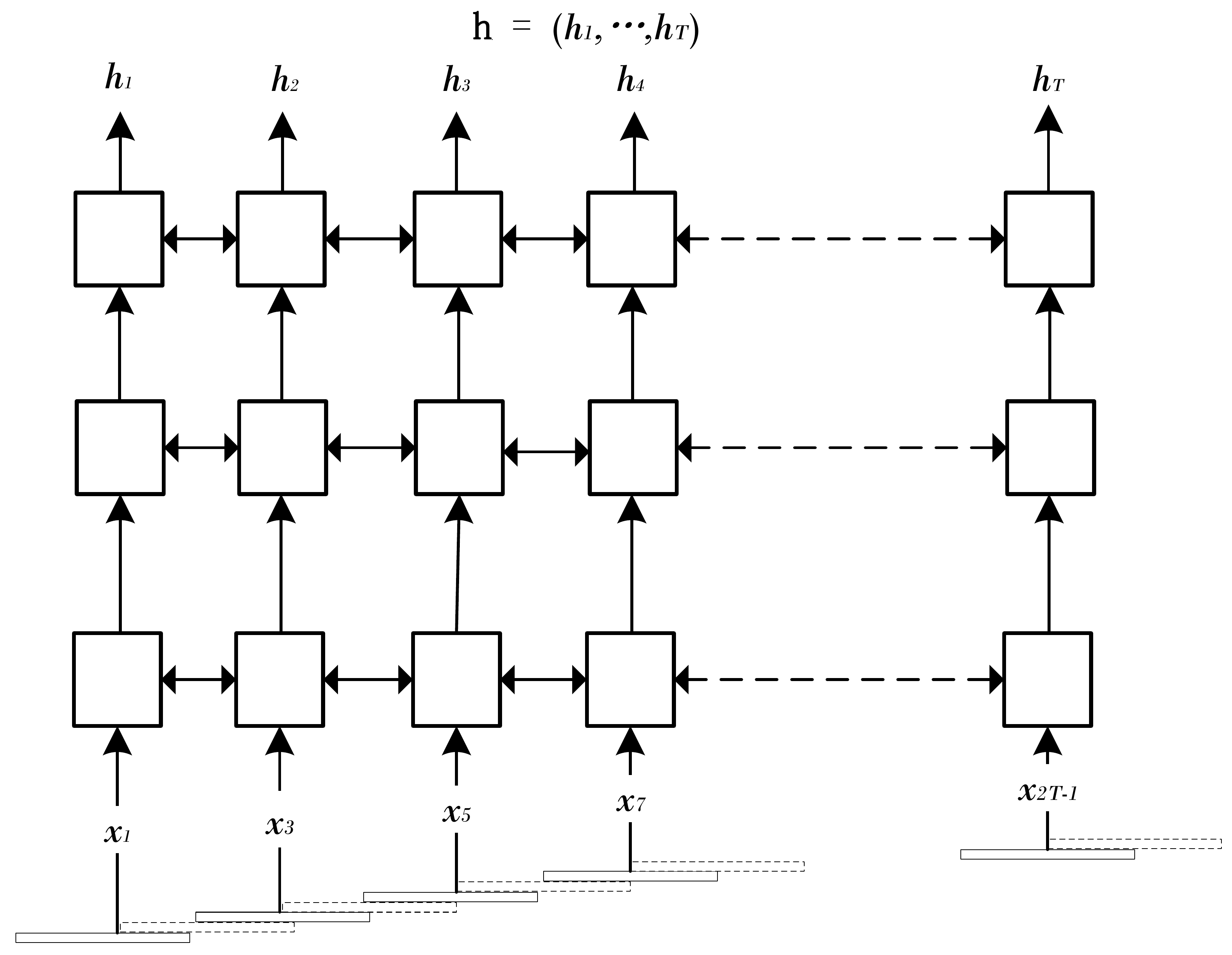}
    	\caption{{\it The encoder model is a BLSTM that extracts $\mathbf{h}$ from input $\mathbf{x}$. Frame skipping is employed during training.}}\vspace{-5pt}
    	\label{fig:encoder}
    \end{figure}

\begin{figure}[t]
    	\centering
    	\includegraphics[width=\columnwidth]{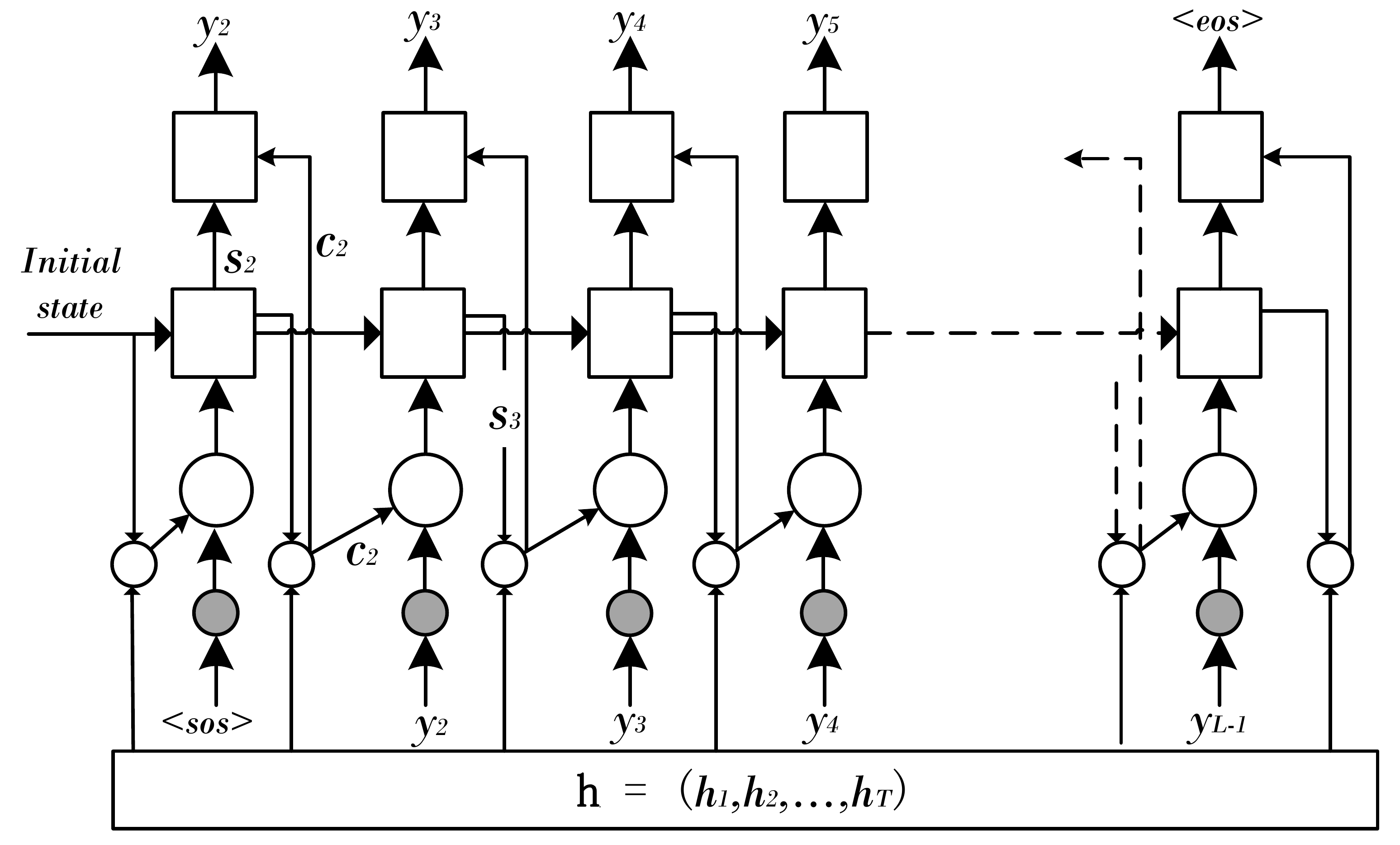}
    	\caption{{\it The AttendAndSpell model composed by MLP (the Attention mechanism) and LSTM (the Decoder model). }}\vspace{-5pt}
    	\label{fig:decoder}
    \end{figure}

    Specifically, in Fig.~\ref{fig:encoder}, the encoder we used is a bidirectional long short term memory (BLSTM) recurrent neural network (RNN) that generates a high-level feature representation sequence $\mathbf{h} = (h_1,...,h_T)$ from the input time-frequency representation sequence of speech $\mathbf{x}$:
    \begin{flalign}
      \mathbf{h}&=Listen(\mathbf{x}) 
    \end{flalign}
    In Fig.~\ref{fig:decoder}, the $AttendAndSpell$ is an attention-based transducer:
     \begin{equation}
      p(\mathbf{y}|\mathbf{x})=AttendAndSpell(\mathbf{y,h}).
     \end{equation}
   In practice, the process predicts the character $y_i$ at a time according to the probability distribution:
   \begin{equation}
      p(y_i|\mathbf{x},y_{i-1},\cdots,y_1)=CharacterDist(s_i,c_i),
     \end{equation}
where $s_i$ is an LSTM hidden state for time $i$, computed by
    \begin{equation}
    s_i=DecodeRNN([y_{i-1},c_{i-1}],s_{i-1}),
    \end{equation}
    and the context vector
    \begin{equation}
        c_i=AttentionContext(s_i,\mathbf{h}).
    \end{equation}
    The $DecodeRNN$ is a unidirectional LSTM RNN which produces a transducer state $s_i$ and the $AttentionContext$ generates context $c_i$ with a multi-layer perceptron (MLP) attention network. Finally, the probability distribution $CharacterDist$ is computed by a softmax function.
    

  \section{Methods}
  In this section, we detail the tricks we used in LAS-based speech recognition for the Mandarin voice search task.
    \subsection{Embedding and regularization}
    Chinese has a large set of characters and even the number of frequently-used characters can reach 3,500. 
    Chan \textit{et. al.}~\cite{chan2016online} have pointed out that the large vocabulary with limited training data made the model difficult to learn and generalize well. This means that, for an end-to-end Mandarin system that directly outputs characters, it is critical to use an appropriate embedding to ensure the converge of the model. 
    
    In this paper, we first represent each character in a one-hot scheme and further embed it to a vector using neural network. Specifically, in Fig.~2, a fully-connected embedding layer (shaded circle) is used to connect the one-hot input and the subsequent BLSTM layer of the LAS encoder. The weight matrix $\mathbf{W}_e$ of the character embedding layer is updated in the whole LAS model training procedure. The embedding layer works as follows. Assume the size of the vocabulary is $n$ and the dimension of the embedding layer is $m$. Then the weight matrix $\mathbf{W}_e$ is of size $n \times m$. When the character's index is $i$, the embedding layer will pass the $i$th row of $\mathbf{W}_e$ to the subsequent encoder. That is, it acts as a lookup-table, making the training procedure more efficient. 
    We find that this simple character embedding provides significant benefit to the model convergence and robustness. 

The LAS model often gives poor generalization to new data without regularizations. Thus two popular regularization tricks are used in this paper: L2 regularization and Gaussian weight noise~\cite{hinton1993keeping, jim1996analysis}.
    
        \subsection{Attention mechanism}
    The attention mechanism selects (or weights) the input frames to generate the next output element. In this study, we compared the content-based attention and the location-based attention.
    
    \textbf{Content-based attention}: Borrowed from neural machine translation~\cite{bahdanau2014neural}, content-based attention can be directly used in speech recognition. Here, the context vector $c_i$ is computed as a weighted sum of $h_i$:
    \begin{equation}
      c_i=\sum_{j=1}^T\alpha_{i,j}h_j.
     \end{equation}
 The weight $\alpha_{i,j}$ of each $h_j$ is computed by
   \begin{equation}
      \alpha_{i,j}=\left.{exp(e_{i,j})} / \right.{\sum_{j=1}^Texp(e_{i,j})}, \label{softmax}
    \end{equation}
where
        \begin{equation}
       e_{i,j}=Score(s_{i-1},h_j).
    \end{equation}
    Here the $Score$ is an MLP network which measures how well the inputs around position $j$ and the output at position
$i$ match. It is based on the LSTM hidden state $s_{i-1}$ and $h_j$ of the input sentence. Specifically, it can be further described by
\begin{equation}
      e_{i,j}=\mathbf{w}^\top tanh(\mathbf{W}\mathbf{s}_{i-1}+\mathbf{V}\mathbf{h}_j+\mathbf{b}),
    \end{equation}
where $\mathbf{w}$ and $\mathbf{b}$ are vectors, and $\mathbf{W}$ and $\mathbf{V}$ are matrices.

    \textbf{Location-based attention}:  In ~\cite{chorowski2015attention}, location-awareness was added to the attention mechanism to better fit the speech recognition task. Specifically, the content-based attention mechanism is extended by making it take into account the alignment at the previous step. $k$ vectors $\mathbf{f}_{i,j}$ are extracted for every position $j$ of the previous alignment $\bm\alpha_{i-1}$ by convolving it with a matrix $\mathbf{F}$:
    \begin{equation}
     \mathbf{f}_{i}=\mathbf{F}*\bm{\alpha}_{i-1} .
    \end{equation}
    By adding $\mathbf{f}_{ij}$, the scoring mechanism is changed to
    \begin{equation}
      e_{i,j}=\mathbf{w}^\top tanh(\mathbf{W}\mathbf{s}_{i-1}+\mathbf{V}\mathbf{h}_j+\mathbf{Uf}_{ij}+\mathbf{b}).
    \end{equation}

    \subsection{Attention smoothing}
    We found that long context information is important for the voice search task. Hence we explore attention smoothing to get longer context in the attention mechanism.
    When the input sequence $\mathbf{h}$ is long, the $\alpha_{i}$ distribution is typically very sharp on convergence, and thus it focuses on only a few frames of $\mathbf{h}$. To keep the diversity of the model, similar to~\cite{chorowski2015attention}, we replace the softmax function in Eq.~(\ref{softmax}) with the logistic sigmoid $\sigma$:
    \begin{equation}
      \alpha_{i,j} = \sigma(e_{i,j}).
    \end{equation}

    \subsection{Frame skipping}
    Frame skipping is a simple-but-effective trick that has been previously used for fast model training and decoding~\cite{miao2016simplifying}. As training BLSTM is notoriously time-consuming, we borrow this idea in the training of LAS encoder which is BLSTM. As our task does not consider online decoding, we use all frames to generate the context $\mathbf{h}$ during decoding.
%
%
%

    \subsection{Language model}
    At each time step, the decoder generates a character depending on the previous ones, similar to the mechanism of a language model (LM). Therefore, the attention model works pretty good without using any explicit language model. However, the model itself is insufficient to learn a complex language model~\cite{bahdanau2016end}.
     Hence we build a character-level language model $T$ from a word-level language model $G$ that is trained using the training transcripts and a lexicon $L$ that simply spells out the characters of each word. In other words, the input of $L$ is characters and output is words. More specifically, we build a finite state transducer (FST) $T = min(det(L \circ G))$ to calculate the log-probability for the character sequences. We add $T$ to the cost of decoder's output:
    \begin{equation}
    C  = -\sum_i[\log p(y_i|\mathbf{x},y_{i-1},\cdots,y_1) + \gamma T]
    \end{equation}
    During decoding, we minimize the cost $C$ which combines the attention-based model and the external language model with a tunable parameter $\gamma$.

    
    
    \section{Experiments}
    
    \subsection{Data}
    
    We used a 3000-hour dataset for LAS model training, which contains approximately 4M voice search utterances, collected from the microphone on the MiTV remote controller. The dataset was composed of diverse search entries on popular TV programs, movies, songs and personal names (e.g. movie stars).
    The test set and held-out validation set were also from the MiTV voice search and each was composed of 3,000 utterances. As input features, we used 80 Mel-scale filterbank coefficients computed every 10ms with delta and delta-delta acceleration coefficients. Mean and variance normalization was conducted for each speaker. For the decoder model, we used 6,925 labels: 6,922 common Chinese characters, unknown token and sentence start and end tokens ($<$sos$>$/$<$eos$>$).

    \subsection{Training}
    We trained LAS models, in which the encoder was a 3-layer BLSTM with 256 LSTM units per-direction (or 512 in total) and the decoder was a 1-layer LSTM with 256 LSTM units. All the weight matrices were initialized with the normalized initialization~\cite{glorot2010understanding} and the bias vectors were initialized to 0. Gradient norm clipping to 1 was applied, together with Gaussian weight noise and L2 weight decay 1e-5. We used ADAM as the optimization method~\cite{kingma2014adam} while we decayed the learning rate from 1e-3 to 1e-4 after it converged. The softmax output and the cross entropy cost were combined as the model cost. Frame skipping was used in the encoder during training. For comparison, we also constructed a CTC model that has the same structure with the the LAS encoder. 
    \subsection{Decoding}
    We used a simple left-to-right beam search algorithm during decoding~\cite{chorowski2015attention}.  
    We invesitaged the importance of the beam-search width on decoding accuracy~\cite{chorowski2015attention} and the impact of the temperature of the softmax function~\cite{chan2016online}. The temperature can smooth the distribution of characters. We changed the character probability distribution by a temperature hyperparameter $\tau$:
    \begin{equation}
    y_t=exp(o_t/\tau) / \sum_j exp(o_j/\tau).
    \end{equation}
where $o_t$ is the input of the softmax function.

    
    \subsection{Results}
    
    \begin{table}[t]
    	\caption{\label{tab:results} {\it Results of our attention-based models with a beam size of 30, $\tau = 2$ and $\gamma=0.1$.}}
    	\vspace{2mm}
    	\footnotesize
    	\centerline{
    		\begin{tabular}{ c  c  c }      			
    			\hline
    			model & CER/\% & SER/\% \\
    			\hline \hline
    			$CTC                        $ & $5.29$ &   $14.57$ \\
    			$Content~based~attention    $ & $4.05$ &   $9.10 $ \\ 
    			$~~+trigram~LM~             $ & $3.60$ &   $7.20 $ \\      			    		
    			$Location~based~attention    $ & $3.82$ &   $8.17 $ \\
    			$~~+trigram~LM              $ & $3.26$ &   $6.33 $ \\
    			$Attention~smoothing        $ & $\textbf{3.58}$ &   $\textbf{7.43}$ \\
    			$~~+trigram~LM              $ & $\textbf{2.81}$ &   $\textbf{5.77}$ \\
    			\hline
    		\end{tabular}
    	}
    \end{table}
    \begin{figure}[t]
    	\centering
    	\includegraphics[width=\columnwidth]{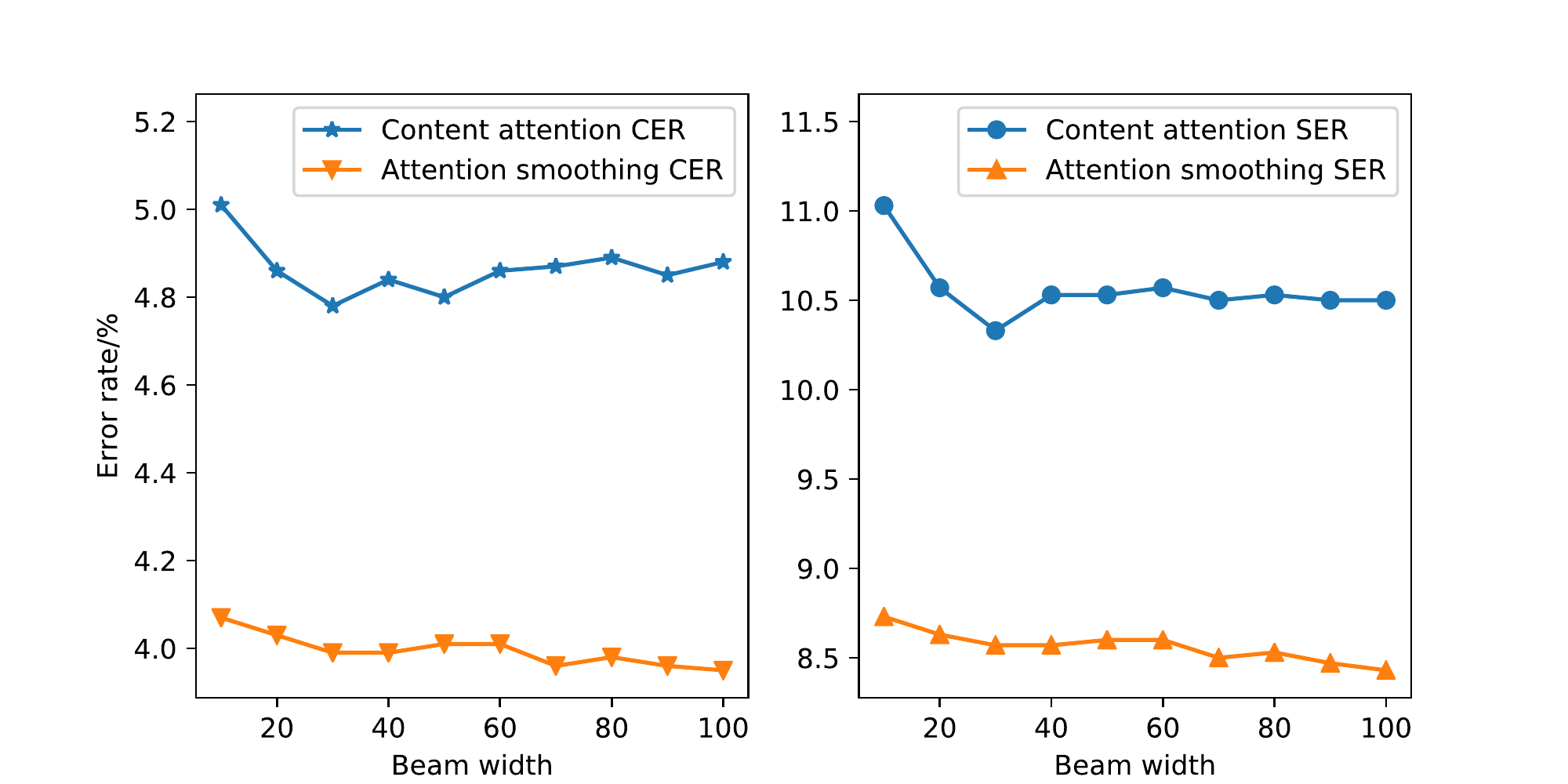}\vspace{-5pt}
    	\caption{{\it The effect of the decoding beam width for the content-based attention and attention smoothing ($\tau = 1$). }}\vspace{-10pt}
    	\label{fig:beam}
    \end{figure}
    
    Table \ref{tab:results} shows that our models performed extremely well in the Mandarin voice search task. The content-based attention model achieved a CER of 4.05\% and a SER of 9.1\%. The location-based attention model achieved a CER of 3.82\% and a SER of 8.17\%, which outperformed the content-based attention model. By using attention smoothing on the content-based attention model, the CER was reduced to 3.58\% ( or 11.6\% relative gain over the content-based attention). We believe that the improvement is mainly because the sigmoid function keeps the diversity of the model and smooths the focus found by the attention mechanism. 
    
    \begin{figure}[t]
    	\centering
    	\includegraphics[width=\columnwidth]{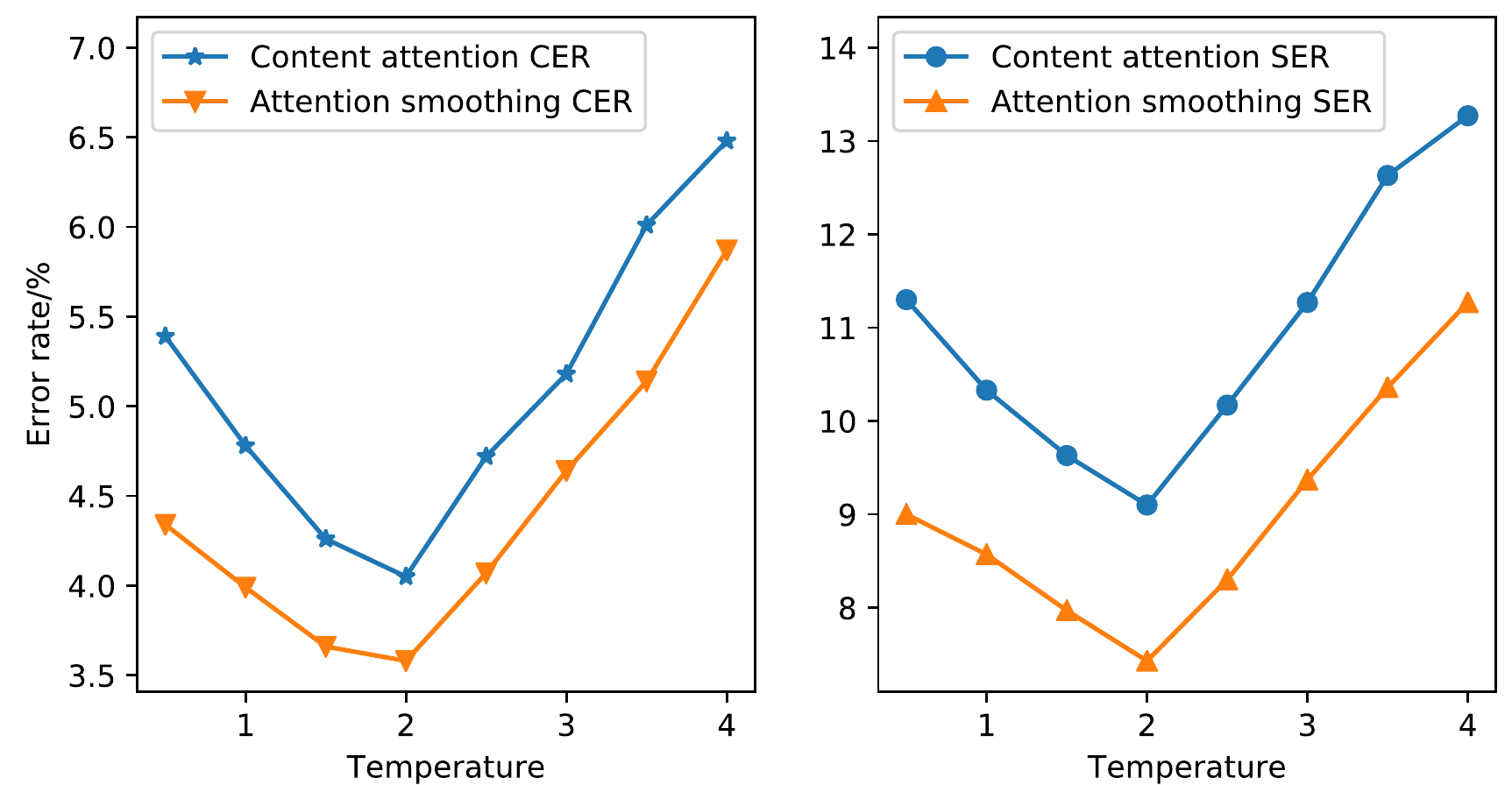}\vspace{-5pt}
    	\caption{{\it The impact of the temperature for content-based attention and attention smoothing (beam-size=30).}} 
    	\label{fig:temp}\vspace{-10pt}
    \end{figure}
    
    Fig.~\ref{fig:beam} shows the effect of the decoding beam width on the WER/SER for the test set. The CER reached the lowest (4.78\%) at a beam width of 30. We cannot observe extra benefit when further increasing the beam width. 
    In Fig.~\ref{fig:temp}, we can see that attention smoothing achieves the best performance when $\tau = 2$ and there is no additional benefits when we further increase the temperature. We see the same observation on the validation set as well.
    Meanwhile, we investigated the effect of adding language model. During decoding, with the help of a trigram LM that was trained using 4M voice search entries, further gains can be observed. Finally, $attention ~smoothing + trigram~LM$ achieved the lowest CER of 2.81\%. This was obtained when $\gamma=0.1$.
    
    \vspace{-10pt}
    \section{Conclusion}\vspace{-5pt}
    In this paper, we reported our preliminary results on attention-based encoder-decoder for Mandarin speech recognition. With several tricks, our model finally achieves a CER of 3.58\% and a SER of 7.43\% on a Mandarin voice search task without a language model. Note that the voice search content on MiTV is kind of limited with closed domains. In the future, we will further investigage our approach on general ASR tasks through public datasets.

    
    \vspace{-10pt}
    \section{Acknowledgments}
    The authors would like to thank the Xiaomi Deep Learning Team and MiAI SRE Team for Xiaomi Cloud-ML and GPU cluster support. We also thank Yu Zhang and Jian Li for helpful comments and suggestions. 
    
    

\bibliographystyle{IEEEtran}
\bibliography{sch-2017}

\end{document}